\documentclass[10pt,twocolumn,letterpaper]{article}

\usepackage[pagenumbers]{cvpr} 

\usepackage{graphicx}
\usepackage{amsmath}
\usepackage{amssymb}
\usepackage{booktabs}
\usepackage{multirow}
\usepackage{algorithm}
\usepackage{algpseudocode}

\usepackage[pagebackref,breaklinks,colorlinks]{hyperref}

\usepackage[capitalize]{cleveref}
\crefname{section}{Sec.}{Secs.}
\Crefname{section}{Section}{Sections}
\Crefname{table}{Table}{Tables}
\crefname{table}{Tab.}{Tabs.}

\def\cvprPaperID{11219} 
\def\confName{CVPR}
\def\confYear{2023}

\begin{document}

\title{Transferable Adversarial Attacks on Vision Transformers with Token Gradient Regularization}
\author{Jianping Zhang$ ^{1} $ \qquad
Yizhan Huang$ ^{1} $ \qquad
Weibin Wu$ ^{2} $\thanks{Corresponding author.} \qquad
Michael R. Lyu$ ^{1} $
\\
$ ^{1} $Department of Computer Science and Engineering, The Chinese University of Hong Kong
\\
$ ^{2} $School of Software Engineering, Sun Yat-sen University
\\
{\tt\small \{jpzhang, yzhuang22, lyu\}@cse.cuhk.edu.hk, wuwb36@mail.sysu.edu.cn}
}

\maketitle

\begin{abstract}
   Vision transformers (ViTs) have been successfully deployed in a variety of computer vision tasks, but they are still vulnerable to adversarial samples. Transfer-based attacks use a local model to generate adversarial samples and directly transfer them to attack a target black-box model. The high efficiency of transfer-based attacks makes it a severe security threat to ViT-based applications. Therefore, it is vital to design effective transfer-based attacks to identify the deficiencies of ViTs beforehand in security-sensitive scenarios. Existing efforts generally focus on regularizing the input gradients to stabilize the updated direction of adversarial samples. However, the variance of the back-propagated gradients in intermediate blocks of ViTs may still be large, which may make the generated adversarial samples focus on some model-specific features and get stuck in poor local optima. To overcome the shortcomings of existing approaches, we propose the Token Gradient Regularization (TGR) method. According to the structural characteristics of ViTs, TGR reduces the variance of the back-propagated gradient in each internal block of ViTs in a token-wise manner and utilizes the regularized gradient to generate adversarial samples. Extensive experiments on attacking both ViTs and CNNs confirm the superiority of our approach. Notably, compared to the state-of-the-art transfer-based attacks, our TGR offers a performance improvement of 8.8\% on average. Code is available at \url{https://github.com/jpzhang1810/TGR}.

\end{abstract}

\section{Introduction}

Transformers have been widely deployed in the natural language processing, achieving state-of-the-art performance. Vision transformer (ViT) \cite{dosovitskiy2020image} first adapts the transformer structure to the computer vision, and manifests excellent performance. Afterward, diverse variants of ViTs have been proposed to further improve its performance \cite{chen2021visformer,touvron2021going} and broaden its application to different computer vision tasks \cite{zheng2021rethinking,zhang2021vit}, which makes ViTs a well-recognized successor for convolutional neural networks (CNNs). Unfortunately, recent studies have shown that ViTs are still vulnerable to adversarial attacks \cite{shao2021adversarial, bhojanapalli2021understanding}, which add human-imperceptible noise to a clean image to mislead deep learning models. It is thus of great importance to understand DNNs \cite{wu2020towards, Wang2020RethinkingTV, huang2022aeon, Wang2022UnderstandingAI} and devise effective attacking methods to identify their deficiencies before deploying them in safety-critical applications \cite{liu2022towards, liu2022surrogate, wu2019deep}.



\begin{figure}[t]
  \centering
  \centerline{\includegraphics[width=0.98\linewidth]{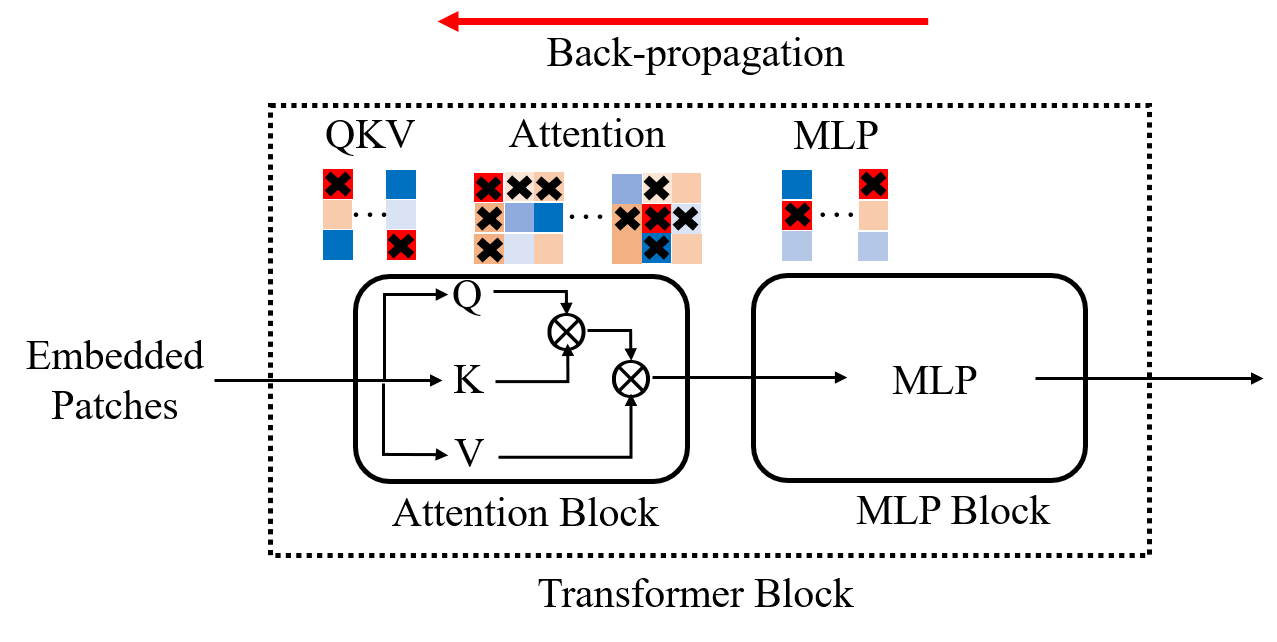}}
  \caption{Illustration of our Token Gradient Regularization (TGR) method. The red-colored entry represents the back-propagated gradient with extreme values. The back-propagated gradients corresponding to one token in the internal blocks of ViTs are called the token gradients. Since we regularize the back-propagated gradients in a token-wise manner,  we eliminate the token gradients (marked with crosses) where extreme gradients locate during back-propagation to reduce the gradient variance. We then use the regularized gradients to generate adversarial samples.}
  \label{fig1}
\end{figure}


Adversarial attacks can be generally partitioned into two categories. The first category is the white-box attack, where attackers can obtain the structures and weights of the target models for generating adversarial samples. The second one is the black-box attack, where attackers cannot fetch the information of the target model. Among different black-box attacks, transfer-based methods employ white-box attacks to attack a local source model and directly transfer the generated adversarial sample to attack the target black-box model. Due to their high efficiency and applicability, transfer-based attacks pose a serious threat to the security of ViT-based applications in practice. Therefore, in this work, we focus on transfer-based attacks on ViTs.

There are generally two branches of transfer-based attacks in the literature \cite{lin2019nesterov}. The first one is based on input transformation, which aims to combine the input gradients of multiple transformed images to generate transferable perturbations. Complementary to such methods, the second branch is based on gradient regularization, which modifies the back-propagated gradient to stabilize the update direction of adversarial samples and escape from poor local optima. For example, Variance Tuning Method (VMI) \cite{wang2021enhancing} tunes the input gradient to reduce the variance of input gradients. However, the variance of the back-propagated gradients in intermediate blocks of ViTs may still be large, which may make the generated adversarial samples focus on some model-specific features with extreme gradient values. As a result, the generated adversarial samples may still get stuck in poor local optima and possess limited transferability across different models.



To address the weaknesses of existing gradient regularization-based approaches, we propose the Token Gradient Regularization (TGR) method for transferable adversarial attacks on ViTs. According to the architecture of ViTs, TGR reduces the variance of the back-propagated gradient in each internal block of ViTs and utilizes the regularized gradient to generate adversarial samples.

More specifically, ViTs crop one image into small patches and treat these patches as a sequence of input tokens to fit the architecture of the transformer. The output tokens of internal blocks in ViTs correspond to the extracted intermediate features. Therefore, we view token representations as basic feature units in ViTs. We then examine the back-propagated gradients of the classification loss with respect to token representations in each internal block of ViTs, which we call token gradient in this work. As illustrated in Figure \ref{fig1}, we directly eliminate the back-propagated gradients with extreme values in a token-wise manner until obtaining the regularized input gradients, which we used to update the adversarial samples. Consequently, we can reduce the variance of the back-propagated gradients in intermediate blocks of ViTs and produce more transferable adversarial perturbations. 

We conducted extensive experiments on the ImageNet dataset to validate the effectiveness of our proposed attack method. We examined the transferability of our generated adversarial samples to different ViTs and CNNs. Notably, compared with the state-of-the-art benchmarks, our proposed TGR shows a significant performance improvement of 8.8\% on average.

We summarize the contributions of this work as below:
\begin{itemize}
    \item We propose the Token Gradient Regularization (TGR) method for transferable adversarial attacks on ViTs. According to the architectures of ViTs, TGR regularizes the back-propagated gradient in each internal block of ViTs in a token-wise manner and utilizes the regularized gradient to generate adversarial samples.

    \item We conducted extensive experiments to validate the effectiveness of our approach. Experimental results confirm that, on average, our approach can outperform the state-of-the-art attacking method with a significant margin of 8.8\% on attacking ViT models and 6.2\% on attacking CNN models.

    \item We showed that our method can be combined with other compatible attack algorithms to further enhance the transferability of the generated adversarial samples. Our method can also be extended to use CNNs as the local source models. 
\end{itemize}

\section{Related Work}

\subsection{Adversarial Attacks on CNNs}

There are generally two categories of adversarial attacks in the literature \cite{lin2019nesterov}: white-box and black-box attacks. White-box attacks get full access to the information of target models, like model structure and weights. In contrast, black-box attacks \cite{wang2022triangle, zhang2023improving} fail to obtain the specifics of target models. White-box adversarial attacks can directly adopt the gradient information of the target model to craft adversarial samples, like the Fast Gradient Sign Method (FGSM) \cite{goodfellow2014explaining} and Basic Iterative Method (BIM) \cite{kurakin2016adversarial}. 

However, in reality, the model structure and weights are hidden from the users. Therefore, more research focuses on the adversarial attack under the black-box setting. Among different black-box attack methodologies, transfer-based attacks stand out due to their severe security threat to deep learning-based applications in practice. Therefore, we also focus on transfer-based attacks in this paper. The ability of adversarial samples crafted by a local source model being able to mislead other black-box target models is called transferability. Transfer-based attacks \cite{zhang2022improving, wu2020boosting, wu2021improving} utilize the transferability of adversarial samples. Specifically, transfer-based attacks usually craft adversarial samples with a local source model using white-box attack methods and input the generated adversarial samples to the target model to cause misclassification. 

There are generally two branches of approaches to enhance the transferability of adversarial samples. The first branch is based on input transformation, aiming to combine the gradient of several transformed images during the generation of transferable perturbations. Diverse Input Method (DIM) \cite{xie2019improving} applies random resizing and padding to the input image with a fixed probability and uses the transformed image to compute the input gradient for updating adversarial samples. Translation Invariant Method (TIM) \cite{dong2019evading} employs shifted images for computing the input gradient, which is simplified by convolving the input gradient of the original image with a kernel matrix. Scale Invariant Method (SIM) \cite{lin2019nesterov} utilizes the scale-invariant property of CNNs and combines the gradients of the scaled copies of the original image together. However, due to the structural difference between ViTs and CNNs, the input transformation methods tailored for CNNs cannot achieve comparable performance improvement for transferable adversarial attacks on ViTs  \cite{wei2022towards}.  Therefore, in this work, we take the special design of ViTs into consideration to improve the transferability of adversarial samples on ViTs.

The second branch is based on gradient regularization \cite{wang2021boosting}, which aims to stabilize the update direction of adversarial samples and escape from the poor local optima. Momentum Iterative Method (MIM) \cite{dong2018boosting} incorporates the momentum term into the computation of the input gradient. Skip Gradient Method (SGM) \cite{wu2020skip} utilizes a decay factor to reduce the back-propagated gradients from the residual module to focus on the transferable low-level information.  Variance Tuning Method (VMI) \cite{wang2021enhancing} tunes the input gradient with the gradient variance in the neighborhood of the target image to reduce the gradient variance. Different from the existing work, we reduce the gradient variance in each internal block of ViTs based on the architecture of ViTs, which avoids the over-reliance on model-specific features and thus improves adversarial transferability against ViTs.

\subsection{Vision Transformer}

ViT \cite{dosovitskiy2020image} first adapts the transformer network structure from the natural language processing field to the computer vision field. Specifically, ViT divides the input image into a sequence of small image patches, which are attached with a classification token to constitute the input to the transformer. More advanced versions of ViTs are proposed to improve the accuracy and efficiency of the vanilla ViT. For example, pooling-based vision transformer (PiT) \cite{heo2021rethinking} decreases the spatial dimension and increases the channel dimension with pooling to improve model capability. Class-attention in image transformer (CaiT) \cite{touvron2021going} builds deeper transformers and adds the classification token in the latter layer of the network. The vision-friendly transformer (Visformer) \cite{chen2021visformer} transit a transformer-based model to a convolution-based model. There are also other works trying to improve the performance of the ViT from other points of view \cite{touvron2021training, han2021transformer}.

\subsection{Adversarial Attacks on ViTs}

With the wide deployment of ViTs in diverse vision tasks, there is an increasing interest in evaluating the robustness of ViTs to identify their deficiencies. Bhojanapalli et al. \cite{bhojanapalli2021understanding}, and Shao et al. \cite{shao2021adversarial} analyzed the robustness of ViTs against white-box attacks. They found that ViTs tend to be more robust than CNNs. Mahmood et al. \cite{mahmood2021robustness} examined the transferability of adversarial samples crafted by ViTs. They discovered that the transferability from ViTs to CNNs is relatively low due to their structural difference. In this work, to better assess the robustness of ViTs in black-box settings, we attempt to improve the transferability of adversarial samples from ViTs to both different ViT models and CNN models.

Existing transfer-based attacks on ViTs generally follow similar methodologies of those on CNNs. For example, the Pay No Attention (PNA) method adapts the SGM to ViTs. Specifically, PNA skips the gradient of the attention block during back-propagation to improve the transferability of adversarial samples, which accounts for a gradient regularization-based approach \cite{wei2022towards}. The PatchOut attack strategy randomly samples a subset of patches to compute the gradient in each attack iteration, which acts as an image transformation method for transferable adversarial attacks on ViTs. Different from such methods, our approach proposes to improve transferable adversarial attacks on ViTs from the perspective of gradient variance reduction, which helps to stabilize the update direction of adversarial samples and escape from the poor local optima.


Different from the above line of research, Nasser et al. \cite{naseer2021improving} proposed the Self-Ensemble (SE) method and the Token Refinement module (TR) to improve the transferability of adversarial samples generated from ViTs. SE utilizes the classification token on each layer of ViTs with a shared classification head to perform feature-level attacks. Based on SE, TR further refines the classification token with fine-tuning to improve attack performance. However, the method is time-consuming since it needs to fine-tune the classification token for each source model. Besides, some variants of ViTs do not have or only have a few classification tokens, like Visformer \cite{chen2021visformer} and CaiT \cite{touvron2021going}. Therefore, the SE and TR method only has limited applicability.

\section{Method}

We first set up some notations. We denote the benign image as $x$ with the image size of $H \times W \times C$, where $H$, $W$, and $C$ represent the height, width, and channel number of the image. ViTs divide the image into a sequence of patches $x_p = \{x_p^{1}, x_p^{1}, \cdots, x_p^{N} \}$, where $x_p^{i}$ is the $i$-th patch of the original image. The shape of each patch $x_p^{i}$ is $P \times P \times C$, and $P$ is the patch size. There are in total $N = \frac{H \cdot W}{P^2}$ patches. The corresponding true label of the image is $y$. We represent the output of a DNN classifier by $f(x)$. $J(x,y; f)$ stands for the classification loss function of the classifier $f$, which is usually the cross-entropy loss. Given the target image $x$, adversarial attacks aim to find an adversarial sample $x^{adv}$, which can mislead the classifier, i.e., $f(x^{adv}) \neq f(x)$, while it is human-imperceptible, i.e., satisfying the constraint $\left\| x - x^{adv} \right\|_{p} < \epsilon$. The $\left\| \cdot \right\|_{p}$ represents the $L_p$ norm, and we focus on the $L_\infty$ norm here to align with previous papers \cite{dong2018boosting}.

\subsection{Token Gradient Regularization}

A large gradient variance \cite{wang2021enhancing} triggers an overfitting issue for generating adversarial samples whose update direction is not optimal and easily stuck in the local optimal. Existing efforts focus on regularizing the update gradient on the input and regardless of the gradient variance in intermediate blocks leading to uncontrollable gradient variance in the back-propagation. The motivation behind our method is to regularize the gradient variance in intermediate blocks of ViTs.

In order to reduce the variance of the back-propagated gradients in intermediate blocks of ViTs, we seek to regularize the gradients and consider the structural characteristics of ViTs. Tokens are the fundamental building blocks in the ViTs, and we call the back-propagated gradients corresponding to one token in the internal blocks of ViTs as the \textit{token gradients}. Consequently, regularizing the gradients in the intermediate blocks is to regularize the token gradients. The proposed Token Gradient Regularization (TGR) is to regularize the token gradients in a token-wise manner. We regard that tokens with extreme back-propagated gradient values contribute to the high gradient variance because the extreme back-propagated gradients tend to be model-specific and unstable features \cite{wang2021enhancing}. 
Specifically, if the back-propagated gradient of a token is in the top-$k$ or bottom-$k$ gradient magnitude among all the tokens, then it is  called the \textit{extreme tokens}, where $k$ is a hyper-parameter. Since we regularize the back-propagated gradients in the token-wise, we eliminate the extreme token gradients to reduce the gradient variance during the back-propagation.

To design effective attack methods that cater to ViTs structural characteristics, we analyze the workflow of ViTs and select representative components in intermediate blocks to employ TGR.  ViTs are composed of transformer blocks and each transformer block has an Attention block and a MLP block. The Attention block deploys the self-attention mechanism to compute the Attention between input tokens by Key and Query and multiply the Attention with Value. Therefore, we consider employing TGR on the QKV component and the Attention component for Attention block. The MLP block utilizes a fully-connected layer to aggregate the channel information for all the tokens. Thus, we select the MLP layer to deploy TGR. In the following sections, we illustrate the detailed implementation of TGR.

\subsection{Implementation}

As mentioned in the previous section, we aim to regularize three components in the architecture of ViTs: the Attention and QKV component in the Attention block and the MLP component in the MLP block.

\textbf{Attention Component.} The Attention component utilizes the multi-head self-attention mechanism to compute the relationship between tokens. We suppose there are $M$ self-attention operations in one Attention component. Therefore, the Attention component will output an attention map with the size of $N \times N \times M$ for one image, where $N$ is the number of patch tokens. We regard that the self-attention head computes the relationship between tokens independently. Thus, we rank the backward gradient in each output channel of the Attention component independently. We first localize the extreme tokens on the attention map and denote their positions. Then, we eliminate the gradient entries that lie in the same rows and columns of the extreme gradients at a time.

\textbf{QKV Component.} QKV component computes the Query, Key, and Value for the self-attention mechanism. Suppose the QKV component has $C$ channel, so the size of the QKV component is $N \times C$ for one image. We also regard the channels are independent, and we rank the backward gradient in each input channel of the QKV component independently. We define the tokens with the top-$k$ or bottom-$k$ back-propagated gradient magnitude as extreme tokens. Thus, we eliminate the extreme gradient entries at a time.

\textbf{MLP Block.} MLP block aggregates the information of each token along the channels. We denote the MLP block has $C$ channel, so the size of the MLP block is $N \times C$ for one image. Similar to the QKV component, we prioritize each input channel of the MLP component independently. We also eliminate the extreme gradient entries of the token with the top-$k$ or bottom-$k$ back-propagated gradient magnitude.

The illustration of the TGR on each component is shown in Figure \ref{fig1}. Apart from regularizing the extreme tokens, we also introduce a scaling factor $s$ on each component to reduce the overall gradient variance. Equation \ref{eq} shows the adversarial attack of step $t$. In the equation, $g'$ is the regularized backward gradient on the input. $modules$ represents the model structure, and $Grads$ records the backward gradient in the network.  $TGR(\cdot)$ is the Token Gradient Regularization method, and details are shown in Algorithm \ref{alg1}.

\begin{equation} \label{eq}
\begin{split}
    g' & = TGR(Grads,modules,k,s)\\
    x^{adv}_{t+1} & = x^{adv}_t + \alpha\cdot sgn\{g'\},
\end{split}
\end{equation}
where $\alpha$ is a hyper-parameter to control the step size in update each iteration.
\begin{algorithm}
\caption{Token Gradient Regularization}\label{alg1}
\begin{algorithmic}
\Require network structure $modules$ and gradients $Grads$
\Require scaling factor $s$ and extreme token number $k$
\Ensure the gradient on the input $g'$

\For{$m \ \textbf{in} \ modules$}
\If{$m$ is MLP or KQV }
    \State $Grads[m] \gets Grads[m]*s$
    \State $token \gets extreme(Grads[m], k)$ \Comment{Extreme Tokens on MLP or KQV component}
    \For{$i = 0 \gets 2k-1$}
    \State $Grads[m][token[i],:] = 0$
    \EndFor
\ElsIf{$m$ is Attention}
    \State $Grads[m] \gets Grads[m]*s$
    \State $tokens \gets extreme(Grads[m], k)$ \Comment{Extreme Token Pairs on the Attention Map}
    \For{$i = 0 \gets 2k-1$}
    \State $Grads[m][tokens[i,0],:,:] = 0$
    \State $Grads[m][:,tokens[i,1],:] = 0$
    \EndFor
\EndIf
\EndFor
\end{algorithmic}
\end{algorithm}

\section{Experiments}

In this section, we present extensive experiments to evaluate the effectiveness of our proposed method. We first clarify the setup of the experiments. After that, we illustrate the attacking results of our method against competitive baseline methods under various experimental settings to show the effectiveness of our proposed attack on both ViT and CNN models. Moreover, we analyze the effect of TGR on the gradient reduction in the internal blocks of ViTs and adapt our approach to CNN attacks. Finally, we perform the ablation study on the selection of the components to employ TGR and the selection of the token number.

\begin{table*}
\centering
\setlength{\tabcolsep}{1.0mm}{
\begin{tabular}{|c|c|cccccccc|} 
\hline
Model & Attack & ViT-B/16 & PiT-B & CaiT-S/24 & Visformer-S & DeiT-B & TNT-S & LeViT-256 & ConViT-B \\ 
\hline
\multirow{5}{*}{ViT-B/16} & MIM & 100.0 & 34.5 & 64.1 & 36.5 & 64.3 & 50.2 & 33.8 & 66.0\\
 & VMI & 99.6 & 48.8 & 74.4 & 49.5 & 73.0 & 64.8 & 50.3 & 75.9\\
 & SGM & 100.0 & 36.9 & 77.1 & 40.1 & 77.9 & 61.6 & 40.2 & 78.4\\
 & PNA & 100.0 & 45.2 & 78.6 & 47.7 & 78.6 & 62.8 & 47.1 & 79.5\\
 & TGR & \bf 100.0 & \bf 49.5 & \bf 85.0 & \bf 53.8 & \bf 85.6 & \bf 73.1 & \bf 56.5 & \bf 85.4 \\ 
\hline
\multirow{5}{*}{PiT-B} & MIM & 24.7 & 100.0 & 34.7 & 44.5 & 33.9 & 43.0 & 38.3 & 37.8\\
 & VMI & 38.9 & 99.7 &  51.0 & 56.6 & 50.1 & 57.0 & 52.6 & 51.7\\
 & SGM & 41.8 & 100.0 & 57.3 & 73.9 & 57.9 & 72.6 & 68.1 & 59.9\\
 & PNA & 47.9 & 100.0 & 62.6 & 74.6 & 62.4 & 70.6 & 67.3 & 61.7\\
 & TGR & \bf 60.3 & \bf 100.0 & \bf 80.2 & \bf 87.3 & \bf 78.0 & \bf 87.1 & \bf 81.6 & \bf 76.5\\ 
\hline
\multirow{5}{*}{CaiT-S/24} & MIM & 70.9 & 54.8 & 99.8 & 55.1 & 90.2 & 76.4 & 54.8 & 88.5\\
 & VMI & 76.3 & 63.6 & 98.8 & 67.3 & 88.5 & 82.3 & 67.0 & 88.1\\
 & SGM & 86.0 & 55.8 & 100.0 & 68.2 & 97.7 & 91.1 & 74.9 & 96.7 \\
 & PNA & 82.4 & 60.7 & 99.7 & 67.7 & 95.7 & 86.9 & 67.1 & 94.0\\
 & TGR & \bf 88.2 & \bf 66.1 & \bf 100.0 & \bf 75.4 & \bf 98.8 & \bf 92.8 & \bf 74.7 & \bf 97.9\\ 
\hline
\multirow{5}{*}{Visformer-S} & MIM & 28.1 & 50.4 & 41.0 & 99.9 & 36.9 & 51.9 & 49.4 & 39.6\\
 & VMI & 39.2 & 60.0 & 56.6 & 100.0 & 54.1 & 62.8 & 59.1 & 54.4\\
 & SGM & 18.8 & 41.8 & 34.9 & 100.0 & 31.2 & 52.1 & 52.7 & 29.5\\
 & PNA & 35.4 & 61.5 & 54.7 & 100.0 & 51.0 & 66.3 & 64.5 & 50.7\\
 & TGR & \bf 41.2 & \bf 70.3 & \bf 62.0 & \bf 100.0 & \bf 59.5 & \bf 74.7 & \bf 74.8 & \bf 56.2\\ 
\hline
\end{tabular}}
\vspace{-2mm}
\caption{The attack success rates (\%) against eight models by various transfer-based attacks. The best results are marked in bold.}
\label{table1}
\vspace{-2mm}
\end{table*}

\begin{table*}
\centering
\setlength{\tabcolsep}{1.0mm}{
\begin{tabular}{|c|c|ccccccc|} 
\hline
Model & Attack & Inc-v3 & Inc-v4 & IncRes-v2 & Res-v2 & Inc-v3$_{\text{ens3}}$ & Inc-v3$_{\text{ens4}}$ & IncRes-v2$_{\text{adv}}$ \\ 
\hline
\multirow{5}{*}{ViT-B/16} & MIM & 31.7 & 28.6 & 26.1 & 29.4 & 22.3 & 19.8 & 16.5\\
 & VMI& 43.1 & 41.6 & 37.9 & 42.6 & 31.4 & 30.6 & 25.0 \\
 & SGM & 31.5 & 27.7 & 23.8 & 28.2 & 20.8 & 18.0 & 14.3  \\
 & PNA & 42.7 & 37.5 & 35.3 & 39.5 & 29.0 & 27.3 & 22.6 \\
 & TGR & \bf 47.5 & \bf 42.3 & \bf 37.6 & \bf 43.3 & \bf 31.5 & \bf 30.8 & \bf 25.6\\ 
\hline
\multirow{5}{*}{PiT-B} & MIM & 36.3 & 34.8 & 27.4 & 29.6 & 19.0 & 18.3 & 14.1  \\
 & VMI & 47.3 & 45.4 & 40.7 & 43.4 & 35.9 & 34.4 & 29.7 \\
 & SGM & 50.6 & 45.4 & 38.4 & 41.9 & 25.6 & 20.8 & 16.7 \\
 & PNA & 59.3 & 56.3 & 49.8 & 53.0 & 33.3 & 32.0 & 25.5 \\
 & TGR & \bf 72.1 & \bf 69.8 & \bf 65.1 & \bf 64.8 & \bf 43.6 & \bf 41.5 & \bf 32.8\\ 
\hline
\multirow{5}{*}{CaiT-S/24} & MIM & 48.4 & 42.9 & 39.5 & 43.8 & 30.8 & 27.6 & 23.3  \\
 & VMI & 58.5 & 50.9 & 48.2 & 52.0 & 38.1 & 36.1 & 30.1 \\
 & SGM & 53.5 & 45.9 & 40.2 & 45.9 & 30.8 & 28.5 & 21.0  \\
 & PNA & 57.2 & 51.8 & 47.7 & 51.6 & 38.4 & 36.2 & 30.1 \\
 & TGR & \bf 60.3 & \bf 52.9 & \bf 49.3 & \bf 53.4 & \bf 39.6 & \bf 37.0 & \bf 31.8\\ 
\hline
\multirow{5}{*}{Visformer-S} & MIM & 44.5 & 42.5 & 36.6 & 39.6 & 24.4 & 20.5 & 16.6  \\
 & VMI & 54.6 & 53.2 & 48.5 & 52.2 & 33.0 & 32.0 & 22.2  \\
 & SGM & 43.2 & 41.1 & 29.6 & 35.7 & 16.1 & 13.0 & 8.2 \\
 & PNA & 55.9 & 54.6 & 46.0 & 51.7 & 29.3 & 26.2 & 21.1 \\
 & TGR & \bf 65.9 & \bf 66.8 & \bf 55.3 & \bf 60.9 & \bf 36.0 & \bf 32.5 & \bf 23.3\\ 
\hline
\end{tabular}}
\vspace{-2mm}
\caption{The attack success rates (\%) against seven models by various transfer-based attacks. The best results are marked in bold.}
\label{table2}
\vspace{-2mm}
\end{table*}

\subsection{Experiment Setup}

We follow the protocol of the baseline method \cite{wei2022towards} to set up the experiments for a fair comparison to attack image classification models trained on ImageNet \cite{russakovsky2015imagenet}. ImageNet is also the most widely utilized benchmark task for transfer-based adversarial attacks \cite{wei2022towards, wang2021enhancing}. Here are the details of the experiment setup.

\textbf{Dataset}. We follow the dataset of the baseline method \cite{wei2022towards} by randomly sampling 1000 images of different categories from the ILSVRC 2012 validation set \cite{russakovsky2015imagenet}. We check that all of the attacking models achieve almost 100\% classification success rate in this paper.

\textbf{Models}. We evaluate the transferability of adversarial samples of ViTs under two attacking scenarios. The first one is that the source and target models are both ViT models to validate the transferability across different ViT structures. The other one is that the source model is ViT, but the target models are CNN models to examine the cross-model structure transferability. We choose four representative ViT models as the source models to generate adversarial samples, including ViT-B/16 \cite{dosovitskiy2020image}, PiT-B \cite{heo2021rethinking}, CaiT-S/24 \cite{touvron2021going}, and Visformer-S \cite{chen2021visformer}. In addition to the four source ViT models, the target ViT models contain four more ViTs: DeiT-B \cite{touvron2021training}, TNT-S \cite{han2021transformer}, LeViT-256 \cite{graham2021levit}, and ConViT-B \cite{d2021convit}. We keep the four ViT source models to craft adversarial samples under the second experiment setting. We select both undefended (normally trained) models and defended (adversarial training and advanced defense technique) models as the target CNN models. For undefended models, we use  four representative target models containing Inception-v3 (Inc-v3) \cite{szegedy2016rethinking}, Inception-v4 (Inc-v4) \cite{szegedy2017inception}, Inception-Resnet-v2 (IncRes-v2) \cite{szegedy2017inception} and Resnet-v2-152 (Res-v2) \cite{he2016deep, he2016identity}. For defended models, we consider adversarial training and advanced defense models because adversarial training is a simple but effective technique \cite{madry2017towards, citation-0}, and advanced defense models are robust against black-box adversarial attacks. Three adversarial trained models are selected: an ensemble of three adversarial trained Inception-v3 models (Inc-v3$_{\text{ens3}}$), an ensemble of four adversarial trained Inception-v3 models (Inc-v3$_{\text{ens4}}$), and adversarial trained Inception-Resnet-v2 (IncRes-v2$_{\text{adv}}$).

\textbf{Baseline Methods}. We first choose the advanced gradient iterative-based method MIM \cite{dong2018boosting} as our baseline. In addition, VMI \cite{wang2021enhancing} is another gradient-based baseline method, which utilizes the gradient variance reduction strategy to regularize the update gradient. Furthermore, we also compare our approach with attacking methods using the structure of the ViTs. We compare our method with SGM \cite{wu2020skip}, which utilizes a decay factor to reduce the gradient from the residual module. In order to show our attacking method outperforming state-of-the-art, we select PNA \cite{wei2022towards} as our baseline, which is the current state-of-the-art attacking approach for ViTs. In addition, we compose all the attacking methods with input transformation method (PatchOut) \cite{wei2022towards} for ViT attacks, which is motivated by DIM \cite{xie2019improving} in CNN attack for ViT models. We denote our method with the PatchOut strategy as TGR-P and the baselines with the PatchOut strategy as MIM-P, VMI-P, SGM-P, and PNA-P, respectively.

\textbf{Evaluation Metric}. In the experiments, the evaluation metric is the attack success rate, the ratio of the adversarial samples which successfully mislead the target model among all the generated adversarial samples.

\textbf{Parameter}. For a fair comparison, we follow the parameter setting in \cite{wei2022towards} to set the maximum perturbation to $\epsilon = 16$ and the number of iterations to $T = 10$, so the step length $\alpha = \frac{\epsilon}{T} = 1.6$. As for the decay factor $\mu$, we set $\mu$ to 1.0 for all the baselines because all the baselines utilize the momentum method as the optimizer. We also keep the hyper-parameter setting of all the baselines to conduct experiments. For the PatchOut strategy, we set the number of sampled patches to be $130$ by following PNA \cite{wei2022towards}. All images are resized to $224 \times 224$ to conduct experiments and set the patch size to be $16$ for the inputs of ViTs. Therefore, the total number of tokens is $16\times 16 = 196$ \textit{without} the classification token or $16\times 16 +1 = 197$ \textit{with} the classification token respectively. We set the scaling factor for the Attention component and MLP block to be $0.25$ and the scaling factor for the QKV component to be $0.75$. In addition, we set $k=1$, which means we only consider the tokens with the maximum or the minimum back-propagated gradient magnitude as extreme tokens.

\subsection{Transferability}

In this section, we analyze the performance of our approach against the undefended ViTs, undefended CNNs, and adversarially trained CNN models respectively. Specifically, we attack a given source model and directly test the other different models by crafted adversarial samples, which is the black-box setting. We also test the adversarial samples on the source model itself in a white-box setting.

\begin{table}
\centering
\setlength{\tabcolsep}{1.0mm}{
\begin{tabular}{|c|c|ccc|} 
\hline
Model & Attack & ViTs & CNNs & CNNs-adv \\ 
\hline
\multirow{5}{*}{ViT-B/16} & MIM-P & 61.3 & 31.3 & 21.7 \\
 & VMI-P & 69.1 & 42.8 & 30.9\\
 & SGM-P & 64.8 & 29.2 & 18.9 \\
 & PNA-P & 70.8 & 42.6 & 29.9 \\
 & TGR-P & \bf 76.0 & \bf 46.7 & \bf 33.3\\ 
\hline
\multirow{5}{*}{PiT-B} & MIM-P & 47.3 & 32.5 & 17.5 \\
 & VMI-P & 59.5 & 46.2 & 35.8 \\
 & SGM-P & 70.0 & 45.6 & 21.3 \\
 & PNA-P & 73.1 & 57.8 & 32.7  \\
 & TGR-P & \bf 82.3 & \bf 68.9 & \bf 41.3\\ 
\hline
\multirow{5}{*}{CaiT-S/24} & MIM-P & 70.3 & 44.0 & 29.3  \\
 & VMI-P & 76.8 & 57.8 & 38.4 \\
 & SGM-P & 85.1 & 49.2 & 29.3 \\
 & PNA-P & 81.6 & 56.6 & 39.3\\
 & TGR-P & \bf 88.8 & \bf 60.5 & \bf 40.5\\ 
\hline
\multirow{5}{*}{Visformer-S} & MIM-P & 54.9 & 45.7 & 23.4\\
 & VMI-P & 64.8 & 56.6 & 32.6\\
 & SGM-P & 51.6 & 44.3 & 15.0\\
 & PNA-P & 68.8 & 61.8 & 32.3\\
 & TGR-P & \bf 70.4 & \bf 64.3 & \bf 33.5\\ 
\hline
\end{tabular}}
\vspace{-2mm}
\caption{The average attack success rates (\%) against ViTs, CNNs, and adversarially trained CNNs by various transfer-based attacks with PatchOut strategy. The best results are marked in bold.}
\vspace{-2mm}
\label{table3}
\end{table}

\begin{table}
\centering
\small
\begin{tabular}{|c|cccc|} 
\hline
Methods & Deep & Middle & Shallow & Average\\ 
\hline
MIM & 7.5  &37.6 &70.9 &38.7\\
VMI &  4.0 &19.1 & 34.0& 19.1\\
TGR  & \bf 1.7 & \bf 5.1 & \bf 6.6 & \bf 4.4\\
\hline
\end{tabular}
\vspace{-2mm}
\caption{The average gradient variance of ViT-B/16 by different attacking methods. The best results are marked in bold.}
\vspace{-2mm}
\label{table4}
\end{table}

\begin{table}
\centering
\setlength{\tabcolsep}{1.0mm}{
\begin{tabular}{|c|c|cccc|} 
\hline
Model & Attack & Inc-v3 & IncRes-v2 & Inc-v3$_{\text{ens4}}$ & IncRes-v2$_{\text{adv}}$ \\ 
\hline
\multirow{3}{*}{Res-v2} & MIM & 59.1 & 50.8 & 18.5 & 11.7 \\
 & VMI & 66.3 & 58.6 & \bf 33.4& \bf 21.5\\
 & TGR & \bf 75.5 & \bf 67.6 & 29.7& 17.2\\ 

\hline
\end{tabular}}
\vspace{-2mm}
\caption{The attack success rates (\%) against four CNN models by various transfer-based attacks on Res-v2. The best results are marked in bold.}
\vspace{-2mm}
\label{table5}
\end{table}

We first craft adversarial samples on ViTs and transfer the adversarial samples to other ViT models. As shown in Table \ref{table1}, our approach achieves nearly 100\% white-box attacking success rate. In addition, our method outperforms all the other baselines with a large margin of 8.8\% attacking accuracy, demonstrating the high transferability of adversarial samples generated by our approach. Although VMI and SGM are attacking methods for CNN models, VMI directly regularizes the gradient on the input, and SGM utilizes skip connections, which are largely used in ViTs. VMI and SGM can achieve good attacking performance on ViTs, but they are still inferior to our approach. In addition, VMI also utilizes the idea of gradient variance reduction to generate adversarial samples, but our approach has a higher transferability, which 
reveals that regularizing the gradient variance in intermediate blocks of the model is effective.

Then, we study the performance of our proposed attacking method against undefended CNN models and adversarially trained CNN models to validate the cross-structure transferability. We transfer the generated adversarial samples from the source ViT models to the target CNN models. The attacking performance is summarized in Table \ref{table2}. The transferability of all the attacking methods drops significantly on target CNN models because of the different architectures of ViTs and CNNs. Our approach consistently outperforms all the baselines with a margin of 6.2\% on average, which demonstrates the superiority of cross-structure transferability of our proposed attacking method. The difference in the transferability between the undefended CNN models and adversarially trained CNN models is small, which demonstrates the adversarial samples crafted by the ViTs can figure out similar defects inside undefended or adversarially trained CNN models. Furthermore, our approach achieves 39.3\% attacking accuracy on adversarial trained CNNs by attacking the PiT-B model on average, which shows a serious threat to the CNN defense methods. Therefore, new defense methods are required to defend the transferable adversarial samples crafted by ViT models.

Furthermore, we compose all the attacking methods with the input transformation method in ViT: the PatchOut to further improve transferability as shown in Tabel \ref{table3}. Our approach, combined with the PatchOut strategy, also outperforms all the baseline methods by a considerable margin of 4.9\% on average under the black-box setting, which further demonstrates the superiority of our method.

\subsection{Analysis}

In addition to evaluating the attacking performance compared with baselines, we aim to understand why our proposed method can achieve good performance. Moreover, we also craft adversarial samples on CNNs based on our proposed TRG method on the CNN layer to show the effectiveness of regularizing the gradient variance in intermediate blocks of the network.

We first compute the gradient variance of each block in the ViT-B/16 during the generation of adversarial samples to show the benefit of regularizing the gradient of intermediate blocks in the network. We randomly sample 100 images and compute the average gradient variance of the network. The whole network is divided into three parts -- shallow level (Block 1 - Block 4), middle level (Block 5 - Block 8), and deep level (Block 8 - Block 12), and we compute the average gradient of each level of the network during the adversarial sample generation. As shown in Table \ref{table4}, the deep level has the least average gradient variance, and the shallow level has the largest average gradient variance because the gradient variance of the blocks increases during the back-propagation process. Furthermore, our proposed TGR achieves the least average gradient variance in all the levels compared with the baselines. Although VMI can reduce the gradient variance of each component during iteration, the gradient variance in the network is still too large. Therefore, regularizing the gradient variance in intermediate blocks by our proposed TGR can mitigate the increment of the gradient variance during the back-propagation.

In addition, we also adapt TGR to CNN attacks to show the general effectiveness of our approach. We apply TGR on the gradient of intermediate feature maps in the CNN during back-propagation. We generate adversarial samples by attacking Res-v2 and test the transferability on four CNN models, as shown in Table \ref{table5}. Our proposed TGR outperforms MIM by more than 12.5\% and achieves similar performance with the baseline VMI. The experiment result shows the effectiveness of our approach on attacking CNN models. Although our TGR is not designed for CNN attacks, our approach can reduce the gradient variance in intermediate blocks of the CNNs contributing to competitive performance.

\subsection{Ablation Study}

\begin{table}
\centering
\small
\begin{tabular}{|ccc|ccc|} 
\hline
Attention & QKV & MLP & ViTs & CNNs & CNNs-adv  \\ 
\hline
- & - & - & 56.2 & 29.0 & 19.5 \\
\checkmark & - & - & 67.4 & 38.1 & 25.4  \\
- & \checkmark & - & 64.1 & 33.7 & 23.1\\
- & - & \checkmark & 57.3 & 30.0 & 19.9 \\
\checkmark & \checkmark & - & 69.7 & 40.0 & 27.3 \\
\checkmark & - & \checkmark & 69.3 & 39.4 & 26.6 \\
- & \checkmark & \checkmark & 66.0 & 35.5 & 23.7 \\
\checkmark & \checkmark & \checkmark & 73.6 & 42.7 & 29.3 \\
\hline
\end{tabular}
\vspace{-2mm}
\caption{The average attack success rates (\%) against ViTs, CNNs, and adversarially trained CNNs by various setting of components.}
\vspace{-2mm}
\label{table6}
\end{table}

\begin{figure}[t]
  \centering
  \centerline{\includegraphics[width=0.85\linewidth]{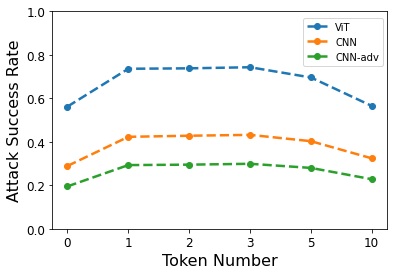}}
  \vspace{-2mm}
  \caption{
  The  attack  success  rates  of  TGR with different number of extreme tokens.
  }
  \vspace{-2mm}
  \label{fig2}
\end{figure}

In this section, we do ablation studies on the influence of two factors on transferability in our proposed TGR: 1) components in ViTs. We want to figure out the contribution of each component to the transferability. 2) The extreme token number $k$. 

\textbf{Components.} We craft adversarial samples by utilizing TGR on different choices of components and observe the transferability. We choose ViT-B/16 as the source model and observe the attacking performance on ViTs, undefended CNNs, and adversarially trained CNNs. As shown in Table \ref{table6}, the Attention component contributes the most transferability. We believe the Attention component computes the relationships between token pairs, which exert a large influence on the output. The ablation study also validates the effectiveness of TGR on different components in ViTs.

\textbf{Token Number.} We measure the transferability of adversarial samples generated from the ViT-B/16 model by altering the extreme token number $k$ in the attacking algorithm  TRG. We observe from Figure \ref{fig2} that when the token number increases from 0 to 1, the transferability boosts. Then the transferability drops after $k = 3$. We regard the performance improvement is due to the regularized tokens, which reduces the gradient variance inside the network. However, the original gradient will be changed largely by regularizing more tokens, contributing to the observed transferability drop. Therefore, in order to balance the performance and the efficiency, we choose $k=1$.

\section{Conclusion}
In this paper, We first analyze the reasons for the low transferability of the gradient regularization-based methods. Although they regularize the gradient variance on the input, the variance in intermediate blocks of the network is still large, and thus models are stuck in local optima. To address the weakness of existing works, we propose the Token Gradient Regularization (TGR) method for transferable attacks. According to the architecture of ViTs, TGR reduces the variance of the back-propagated gradient in each internal block of ViTs and utilizes the regularized gradient to generate adversarial samples. Extensive experiments on attacking both ViTs and CNNs confirm the superiority of our approach. 

\section*{Acknowledgment}
The work described in this paper was supported by the National Natural Science Foundation of China (Grant No. 62206318) and the Research Grants Council of the Hong Kong Special Administrative Region, China (CUHK 14206921 of the General Research Fund).

{\small
\bibliographystyle{ieee_fullname}
\bibliography{egbib}

\begin{thebibliography}{10}\itemsep=-1pt

\bibitem{bhojanapalli2021understanding}
Srinadh Bhojanapalli, Ayan Chakrabarti, Daniel Glasner, Daliang Li, Thomas
  Unterthiner, and Andreas Veit.
\newblock Understanding robustness of transformers for image classification.
\newblock In {\em Proceedings of the IEEE/CVF International Conference on
  Computer Vision}, pages 10231--10241, 2021.

\bibitem{chen2021visformer}
Zhengsu Chen, Lingxi Xie, Jianwei Niu, Xuefeng Liu, Longhui Wei, and Qi Tian.
\newblock Visformer: The vision-friendly transformer.
\newblock In {\em Proceedings of the IEEE/CVF International Conference on
  Computer Vision}, pages 589--598, 2021.

\bibitem{dong2018boosting}
Yinpeng Dong, Fangzhou Liao, Tianyu Pang, Hang Su, Jun Zhu, Xiaolin Hu, and
  Jianguo Li.
\newblock Boosting adversarial attacks with momentum.
\newblock In {\em Proceedings of the IEEE conference on computer vision and
  pattern recognition}, pages 9185--9193, 2018.

\bibitem{dong2019evading}
Yinpeng Dong, Tianyu Pang, Hang Su, and Jun Zhu.
\newblock Evading defenses to transferable adversarial examples by
  translation-invariant attacks.
\newblock In {\em Proceedings of the IEEE/CVF Conference on Computer Vision and
  Pattern Recognition}, pages 4312--4321, 2019.

\bibitem{dosovitskiy2020image}
Alexey Dosovitskiy, Lucas Beyer, Alexander Kolesnikov, Dirk Weissenborn,
  Xiaohua Zhai, Thomas Unterthiner, Mostafa Dehghani, Matthias Minderer, Georg
  Heigold, Sylvain Gelly, et~al.
\newblock An image is worth 16x16 words: Transformers for image recognition at
  scale.
\newblock {\em arXiv preprint arXiv:2010.11929}, 2020.

\bibitem{d2021convit}
St{\'e}phane d’Ascoli, Hugo Touvron, Matthew~L Leavitt, Ari~S Morcos, Giulio
  Biroli, and Levent Sagun.
\newblock Convit: Improving vision transformers with soft convolutional
  inductive biases.
\newblock In {\em International Conference on Machine Learning}, pages
  2286--2296. PMLR, 2021.

\bibitem{goodfellow2014explaining}
Ian~J Goodfellow, Jonathon Shlens, and Christian Szegedy.
\newblock Explaining and harnessing adversarial examples.
\newblock {\em arXiv preprint arXiv:1412.6572}, 2014.

\bibitem{graham2021levit}
Benjamin Graham, Alaaeldin El-Nouby, Hugo Touvron, Pierre Stock, Armand Joulin,
  Herv{\'e} J{\'e}gou, and Matthijs Douze.
\newblock Levit: a vision transformer in convnet's clothing for faster
  inference.
\newblock In {\em Proceedings of the IEEE/CVF international conference on
  computer vision}, pages 12259--12269, 2021.

\bibitem{han2021transformer}
Kai Han, An Xiao, Enhua Wu, Jianyuan Guo, Chunjing Xu, and Yunhe Wang.
\newblock Transformer in transformer.
\newblock {\em Advances in Neural Information Processing Systems},
  34:15908--15919, 2021.

\bibitem{he2016deep}
Kaiming He, Xiangyu Zhang, Shaoqing Ren, and Jian Sun.
\newblock Deep residual learning for image recognition.
\newblock In {\em Proceedings of the IEEE conference on computer vision and
  pattern recognition}, pages 770--778, 2016.

\bibitem{he2016identity}
Kaiming He, Xiangyu Zhang, Shaoqing Ren, and Jian Sun.
\newblock Identity mappings in deep residual networks.
\newblock In {\em European conference on computer vision}, pages 630--645.
  Springer, 2016.

\bibitem{heo2021rethinking}
Byeongho Heo, Sangdoo Yun, Dongyoon Han, Sanghyuk Chun, Junsuk Choe, and
  Seong~Joon Oh.
\newblock Rethinking spatial dimensions of vision transformers.
\newblock In {\em Proceedings of the IEEE/CVF International Conference on
  Computer Vision}, pages 11936--11945, 2021.

\bibitem{huang2022aeon}
Jen-tse Huang, Jianping Zhang, Wenxuan Wang, Pinjia He, Yuxin Su, and Michael~R
  Lyu.
\newblock Aeon: a method for automatic evaluation of nlp test cases.
\newblock In {\em Proceedings of the 31st ACM SIGSOFT International Symposium
  on Software Testing and Analysis}, pages 202--214, 2022.

\bibitem{kurakin2016adversarial}
Alexey Kurakin, Ian Goodfellow, Samy Bengio, et~al.
\newblock Adversarial examples in the physical world, 2016.

\bibitem{lin2019nesterov}
Jiadong Lin, Chuanbiao Song, Kun He, Liwei Wang, and John~E Hopcroft.
\newblock Nesterov accelerated gradient and scale invariance for adversarial
  attacks.
\newblock {\em arXiv preprint arXiv:1908.06281}, 2019.

\bibitem{liu2022towards}
Zihan Liu, Yun Luo, Lirong Wu, Zicheng Liu, and Stan~Z Li.
\newblock Towards reasonable budget allocation in untargeted graph structure
  attacks via gradient debias.
\newblock In {\em Advances in Neural Information Processing Systems}, 2022.

\bibitem{liu2022surrogate}
Zihan Liu, Yun Luo, Zelin Zang, and Stan~Z Li.
\newblock Surrogate representation learning with isometric mapping for gray-box
  graph adversarial attacks.
\newblock In {\em Proceedings of the Fifteenth ACM International Conference on
  Web Search and Data Mining}, pages 591--598, 2022.

\bibitem{madry2017towards}
Aleksander Madry, Aleksandar Makelov, Ludwig Schmidt, Dimitris Tsipras, and
  Adrian Vladu.
\newblock Towards deep learning models resistant to adversarial attacks.
\newblock {\em arXiv preprint arXiv:1706.06083}, 2017.

\bibitem{mahmood2021robustness}
Kaleel Mahmood, Rigel Mahmood, and Marten Van~Dijk.
\newblock On the robustness of vision transformers to adversarial examples.
\newblock In {\em Proceedings of the IEEE/CVF International Conference on
  Computer Vision}, pages 7838--7847, 2021.

\bibitem{naseer2021improving}
Muzammal Naseer, Kanchana Ranasinghe, Salman Khan, Fahad~Shahbaz Khan, and
  Fatih Porikli.
\newblock On improving adversarial transferability of vision transformers.
\newblock {\em arXiv preprint arXiv:2106.04169}, 2021.

\bibitem{russakovsky2015imagenet}
Olga Russakovsky, Jia Deng, Hao Su, Jonathan Krause, Sanjeev Satheesh, Sean Ma,
  Zhiheng Huang, Andrej Karpathy, Aditya Khosla, Michael Bernstein, et~al.
\newblock Imagenet large scale visual recognition challenge.
\newblock {\em International journal of computer vision}, 115(3):211--252,
  2015.

\bibitem{shao2021adversarial}
Rulin Shao, Zhouxing Shi, Jinfeng Yi, Pin-Yu Chen, and Cho-Jui Hsieh.
\newblock On the adversarial robustness of vision transformers.
\newblock {\em arXiv preprint arXiv:2103.15670}, 2021.

\bibitem{szegedy2017inception}
Christian Szegedy, Sergey Ioffe, Vincent Vanhoucke, and Alexander~A Alemi.
\newblock Inception-v4, inception-resnet and the impact of residual connections
  on learning.
\newblock In {\em Thirty-first AAAI conference on artificial intelligence},
  2017.

\bibitem{szegedy2016rethinking}
Christian Szegedy, Vincent Vanhoucke, Sergey Ioffe, Jon Shlens, and Zbigniew
  Wojna.
\newblock Rethinking the inception architecture for computer vision.
\newblock In {\em Proceedings of the IEEE conference on computer vision and
  pattern recognition}, pages 2818--2826, 2016.

\bibitem{touvron2021training}
Hugo Touvron, Matthieu Cord, Matthijs Douze, Francisco Massa, Alexandre
  Sablayrolles, and Herv{\'e} J{\'e}gou.
\newblock Training data-efficient image transformers \& distillation through
  attention.
\newblock In {\em International Conference on Machine Learning}, pages
  10347--10357. PMLR, 2021.

\bibitem{touvron2021going}
Hugo Touvron, Matthieu Cord, Alexandre Sablayrolles, Gabriel Synnaeve, and
  Herv{\'e} J{\'e}gou.
\newblock Going deeper with image transformers.
\newblock In {\em Proceedings of the IEEE/CVF International Conference on
  Computer Vision}, pages 32--42, 2021.

\bibitem{Wang2022UnderstandingAI}
Wenxuan Wang, Wenxiang Jiao, Yongchang Hao, Xing Wang, Shuming Shi, Zhaopeng
  Tu, and Michael~R. Lyu.
\newblock Understanding and improving sequence-to-sequence pretraining for
  neural machine translation.
\newblock In {\em Annual Meeting of the Association for Computational
  Linguistics}, 2022.

\bibitem{Wang2020RethinkingTV}
Wenxuan Wang and Zhaopeng Tu.
\newblock Rethinking the value of transformer components.
\newblock In {\em International Conference on Computational Linguistics}, 2020.

\bibitem{wang2021enhancing}
Xiaosen Wang and Kun He.
\newblock Enhancing the transferability of adversarial attacks through variance
  tuning.
\newblock In {\em Proceedings of the IEEE/CVF Conference on Computer Vision and
  Pattern Recognition}, pages 1924--1933, 2021.

\bibitem{wang2021boosting}
Xiaosen Wang, Jiadong Lin, Han Hu, Jingdong Wang, and Kun He.
\newblock {Boosting Adversarial Transferability through Enhanced Momentum}.
\newblock In {\em British Machine Vision Conference}, 2021.

\bibitem{wang2022triangle}
Xiaosen Wang, Zeliang Zhang, Kangheng Tong, Dihong Gong, Kun He, Zhifeng Li,
  and Wei Liu.
\newblock Triangle attack: A query-efficient decision-based adversarial attack.
\newblock In {\em Computer Vision--ECCV 2022: 17th European Conference, Tel
  Aviv, Israel, October 23--27, 2022, Proceedings, Part V}, pages 156--174.
  Springer, 2022.

\bibitem{wei2022towards}
Zhipeng Wei, Jingjing Chen, Micah Goldblum, Zuxuan Wu, Tom Goldstein, and
  Yu-Gang Jiang.
\newblock Towards transferable adversarial attacks on vision transformers.
\newblock In {\em Proceedings of the AAAI Conference on Artificial
  Intelligence}, volume~36, pages 2668--2676, 2022.

\bibitem{wu2020skip}
Dongxian Wu, Yisen Wang, Shu-Tao Xia, James Bailey, and Xingjun Ma.
\newblock Skip connections matter: On the transferability of adversarial
  examples generated with resnets.
\newblock {\em arXiv preprint arXiv:2002.05990}, 2020.

\bibitem{wu2020boosting}
Weibin Wu, Yuxin Su, Xixian Chen, Shenglin Zhao, Irwin King, Michael~R Lyu, and
  Yu-Wing Tai.
\newblock Boosting the transferability of adversarial samples via attention.
\newblock In {\em Proceedings of the IEEE/CVF Conference on Computer Vision and
  Pattern Recognition}, pages 1161--1170, 2020.

\bibitem{wu2020towards}
Weibin Wu, Yuxin Su, Xixian Chen, Shenglin Zhao, Irwin King, Michael~R Lyu, and
  Yu-Wing Tai.
\newblock Towards global explanations of convolutional neural networks with
  concept attribution.
\newblock In {\em Proceedings of the IEEE/CVF Conference on Computer Vision and
  Pattern Recognition}, pages 8652--8661, 2020.

\bibitem{wu2021improving}
Weibin Wu, Yuxin Su, Michael~R Lyu, and Irwin King.
\newblock Improving the transferability of adversarial samples with adversarial
  transformations.
\newblock In {\em Proceedings of the IEEE/CVF Conference on Computer Vision and
  Pattern Recognition}, pages 9024--9033, 2021.

\bibitem{wu2019deep}
Weibin Wu, Hui Xu, Sanqiang Zhong, Michael~R Lyu, and Irwin King.
\newblock Deep validation: Toward detecting real-world corner cases for deep
  neural networks.
\newblock In {\em 2019 49th Annual IEEE/IFIP International Conference on
  Dependable Systems and Networks (DSN)}, pages 125--137. IEEE, 2019.

\bibitem{xie2019improving}
Cihang Xie, Zhishuai Zhang, Yuyin Zhou, Song Bai, Jianyu Wang, Zhou Ren, and
  Alan~L Yuille.
\newblock Improving transferability of adversarial examples with input
  diversity.
\newblock In {\em Proceedings of the IEEE/CVF Conference on Computer Vision and
  Pattern Recognition}, pages 2730--2739, 2019.

\bibitem{citation-0}
Zhuoer Xu, Guanghui Zhu, Changhua Meng, Zhenzhe Ying, Weiqiang Wang, GU Ming,
  Yihua Huang, et~al.
\newblock A2: Efficient automated attacker for boosting adversarial training.
\newblock In {\em Advances in Neural Information Processing Systems}, 2022.

\bibitem{zhang2022improving}
Jianping Zhang, Weibin Wu, Jen-tse Huang, Yizhan Huang, Wenxuan Wang, Yuxin Su,
  and Michael~R Lyu.
\newblock Improving adversarial transferability via neuron attribution-based
  attacks.
\newblock In {\em Proceedings of the IEEE/CVF Conference on Computer Vision and
  Pattern Recognition}, pages 14993--15002, 2022.

\bibitem{zhang2023improving}
Zeliang Zhang, Peihan Liu, Xiaosen Wang, and Chenliang Xu.
\newblock Improving adversarial transferability with scheduled step size and
  dual example.
\newblock {\em arXiv preprint arXiv:2301.12968}, 2023.

\bibitem{zhang2021vit}
Zixiao Zhang, Xiaoqiang Lu, Guojin Cao, Yuting Yang, Licheng Jiao, and Fang
  Liu.
\newblock Vit-yolo: Transformer-based yolo for object detection.
\newblock In {\em Proceedings of the IEEE/CVF International Conference on
  Computer Vision}, pages 2799--2808, 2021.

\bibitem{zheng2021rethinking}
Sixiao Zheng, Jiachen Lu, Hengshuang Zhao, Xiatian Zhu, Zekun Luo, Yabiao Wang,
  Yanwei Fu, Jianfeng Feng, Tao Xiang, Philip~HS Torr, et~al.
\newblock Rethinking semantic segmentation from a sequence-to-sequence
  perspective with transformers.
\newblock In {\em Proceedings of the IEEE/CVF conference on computer vision and
  pattern recognition}, pages 6881--6890, 2021.

\end{thebibliography}
}

\end{document}